\documentclass[runningheads]{llncs}
\usepackage{times}
\usepackage{latexsym}
\usepackage{pifont}
\usepackage{xcolor}
\newcommand{\etal}{\emph{et al.}}
\usepackage{graphicx}
\usepackage{amsmath,amssymb} 
\usepackage{mathtools}
\usepackage{color}
\usepackage{booktabs}
\usepackage{algorithm}
\usepackage{algpseudocode}
\usepackage[export]{adjustbox} 
\usepackage{array,multirow}
\usepackage{url}
\usepackage{tikz}
\usepackage{cleveref}

\begin{document}
\title{Machine Unlearning for Document Classification}
%
%
%

\author{Lei Kang\inst{1}\orcidID{0000-0002-1962-3916} \and \\
Mohamed Ali	Souibgui\inst{1}\orcidID{0000-0003-0100-9392} \and \\
Fei	Yang\inst{2}\orcidID{0000-0003-4099-6511} \and \\
Lluis Gomez\inst{1}\orcidID{0000-0003-1408-9803} \and \\
Ernest Valveny\inst{1}\orcidID{0000-0002-0368-9697}  \and \\
Dimosthenis Karatzas\inst{1}\orcidID{0000-0001-8762-4454}}
\authorrunning{L. Kang et al.}
%
\institute{Computer Vision Center, Universitat Autònoma de Barcelona, Spain 
\and College of Computer Science, Nankai University, China
\email{\{lkang,msouibgui,lgomez,ernest,dimos\}@cvc.uab.es feiyang@nankai.edu.cn}
}

\maketitle              
\begin{abstract}
Document understanding models have recently demonstrated remarkable performance by leveraging extensive collections of user documents.  However, since documents often contain large amounts of personal data, their usage can pose a threat to user privacy and weaken the bonds of trust between humans and AI services. In response to these concerns, legislation advocating ``the right to be forgotten" has recently been proposed, allowing users to request the removal of private information from computer systems and neural network models. A novel approach, known as machine unlearning, has emerged to make AI models forget about a particular class of data. In our research, we explore machine unlearning for document classification problems, representing, to the best of our knowledge, the first investigation into this area. Specifically, we consider a realistic scenario where a remote server houses a well-trained model and possesses only a small portion of training data. This setup is designed for efficient forgetting manipulation. This work represents a pioneering step towards the development of machine unlearning methods aimed at addressing privacy concerns in document analysis applications. Our code is publicly available at \url{https://github.com/leitro/MachineUnlearning-DocClassification}. 

\keywords{Machine Unlearning  \and Document Classification \and Data Privacy \and Forgetting.}
\end{abstract}

\section{Introduction}

Document classification is a process to assign different categories to documents as required, eventually helping with storage, management and analysis of the documents. It has become an important and basic part of the daily functioning of many institutes and companies nowadays. In document analysis community, towards modern document classification~\cite{harley2015evaluation} and historical document classification~\cite{seuret2021icdar} problems, there are active state-of-the-art methods popping up using visual modality~\cite{kumar2014structural,afzal2015deepdocclassifier}, text modality~\cite{adhikari2019docbert}, and multi-modalities~\cite{bakkali2020cross,bakkali2023vlcdoc}. In industry, business-level document classification service~\footnote{https://help.sap.com/docs/document-classification} has provided user-friendly access for customers to upload their private documents and utilize the high-performance document classification service. This service is easy to use for customers with different background, however, there might be risk of privacy leakage when uploading private documents to the AI service provider for training. 

With the increasing privacy threats in nowadays machine learning models, many concerns are raised that cause governments to pose regularization to protect users' privacy. For example, the General Data Protection Regulation (GDPR) which is set by the European Union. This regulation mandates that corporations adhere to stringent privacy standards, including the prompt deletion of user information upon request \cite{regulation2018general}. Thus, how to efficiently forget the private knowledge in a well-trained document classification model on customers' request, becomes a challenging and urgent need to be solved. 

A naive approach to forgetting specific knowledge is to retrain the model with the whole training set that only excludes the target data to be forgotten. However, this process is costly according to human and computing resources. Thus, a new privacy protection research line has emerged, named Machine Unlearning, aiming to efficiently remove the knowledge of the forgetting data from a well-trained model while maintaining the performance on the retained data. Retraining from scratch is costly but guarantees an exact forgetting on the target data, so it is named as exact unlearning and its performance becomes the upper-bound gold standard. A more realistic and efficient strategy is called approximate unlearning, which relies on the well-trained model to implement minor updates to the model parameters, with the aim of forgetting the target data without inducing catastrophic forgetting on the retained data. Nevertheless, these approximate unlearning methods are difficult to implement due to the lack of interpretability of the well-trained model. Most recent machine unlearning methods~\cite{lin2023erm,chen2023boundary} focus on image classification problems such as CIFAR10~\cite{krizhevsky2009learning}, CIFAR100~\cite{krizhevsky2009learning}, Tiny-ImageNet~\cite{le2015tiny} and Vggface2~\cite{cao2018vggface2}. As far as we know, there is no machine unlearning study on document classification problem. Furthermore, the state-of-the-art machine unlearning methods usually have a strong assumption that there is access to not only the well-trained model but also the full training data, which is not realistic for AI service provider due to the large storage and privacy issues. In this paper, we focus on document classification scenario, conducting a thorough investigation into various machine unlearning strategies within a constraint of limited access to training data. The summarization of our major contributions is as follows:

\begin{itemize}
    \item We propose machine unlearning methods for document classification problem, which is as far as we know, the first work towards forgetting customers' private information on demand in the document analysis community.  
    \item Unlike the state-of-the-art machine unlearning methods that have access to the full training data, we constrain the training data usage to 10\% and even none for the forget set. This constraint makes the study more practical for potential future use cases.
    \item We've developed a label-guided sample generator that creates a synthetic forget set, allowing for unlearning without the necessity of storing the real forget data. Comprehensive experiments are done to validate the effective performance of our proposed methods.
\end{itemize}

\section{Related Work}

Document classification methods have been well studied during recent years. With the proliferation of Optical Character Recognition (OCR) tools and the advancement of Large Language Models (LLMs), document images can now be accurately transcribed into text. Subsequently, utilizing BERT, document classification can be achieved solely through textual modality~\cite{adhikari2019docbert}.  Convolutional Neural Networks (CNNs) models can also deal with the document classification problem using visual modality~\cite{afzal2015deepdocclassifier}. Recently, multi-modality document classifiers have achieved a promising performance using both textual and visual modalities~\cite{bakkali2020cross,bakkali2023vlcdoc}. In this paper, we focus on unlearning process for document classification models, thus we propose to utilize a simple CNNs model as the document classifier with solely visual modality, avoiding confusion caused by multi-modalities. 

The concept of machine unlearning was initially introduced by Cao~\etal~\cite{cao2015towards} as a data forgetting algorithm in statistical query learning. Brophy~\etal~\cite{brophy2021machine} explored data forgetting algorithm with minimal retraining for random forests. Ginart~\etal~\cite{ginart2019making} suggested a data deletion method in k-means clustering. Izzo~\etal~\cite{izzo2021approximate} presented a
projective residual update method to delete data points from linear models. Baumhauer~\etal~\cite{baumhauer2022machine} proposed a method to conceal the class information from the output logits. However, this does not remove the information present in the network weights. Neel~\etal~\cite{neel2021descent} studied the results of gradient descent based approach to unlearning in convex models. 

Graves~\etal~\cite{graves2021amnesiac} leveraged gradients to estimate the contributions of individual data samples of retain and forget categories. They stored gradients of batches consisting of target data points and then subtracted the target gradients to update the model’s parameters. For class-specific unlearning tasks or large-scale unlearning requests, it needs to store gradients of almost every batch, and the model will degenerate to the initialization state. Guo~\etal~\cite{guo2020certified} utilized the Influence Theory~\cite{koh2017understanding} to develop a certified information removal framework based on Newton's update removal mechanism. Golatkar~\etal~\cite{golatkar2020eternal} utilized the Fisher Information~\cite{martens2020new} to approximately remove the information of the deleted training samples from general models. Both methods of Guo~\etal~\cite{guo2020certified} and Golatkar~\etal~\cite{golatkar2020eternal} need to calculate the hessian matrix, which is very expensive in deep learning models. Besides, they may compromise the model’s performance due to imprecise estimates of contributions on CNNs and even cause disastrous forgetting. 

Tarun~\etal~\cite{tarun2023fast} propose a fast yet effective machine unlearning framework employing error-maximizing noise generation and impair-repair based weight manipulation. This framework offers a solution without requiring a real forget set, yet still necessitates full access to the retain set. This is interesting in zero-glance privacy setting, because users have the right to delete the private data immediately and the well-trained model cannot be updated by accessing the forget categories of data for deletion, unlike the typical machine unlearning. Dataset distillation~\cite{wang2018dataset} involves generating a limited set of data points. These points need not adhere strictly to the correct data distribution. However, when utilized as training data for the learning algorithm, they are expected to approximate the model trained on the original dataset. Inspired by both of the works above, we propose a label-guided sample generator that can produce synthetic samples based on the well-trained document classification model with frozen weights. These generated samples can be utilized to replace the real forget set. The proposed sample generator is unlike the typical image synthesizers~\cite{zhu2017unpaired,kang2021content}, because we do not expect the generated samples to be visually appealing as the real ones. Instead, our main purpose is to utilize the generated samples to replace the role of the real forget set during the unlearning process. 

In this paper, we focus on the investigation on unlearning behaviour with limited real training data and limited training iterations, achieving a good trade-off between accuracy and efficiency. Additionally, we investigate the feasibility of employing generated samples to replace the real forget set in the unlearning process.

\section{Unlearning for Document Classification}

\subsection{Problem Formulation}
We assume a practical scenario that users contribute different categories of documents to a AI service provider, so that a well-trained model can be obtained with a good performance for document classification. About the users' documents, once the model is properly trained, only a small portion (10\%) of real data will be stored in the server, considering storage capacity and privacy protection. Then, we allow users to request for deletion of their data from the storage and the knowledge from the well-trained model. Furthermore, we equip our framework capacity to proceed with the immediate removal of requested data while keeping only the retain data to perform unlearning. 

The training data $D = \{x_i, y_i\}^{N}_{i=1}$ where $x$ are the document images, $y \in \{0, ..., K\}$ are the corresponding category labels, $N$ is the total number of the training set, and $K$ is the maximum index number of categories. The training set $D$ is utilized to train a document classification model $\mathcal{M}$. Once the model $\mathcal{M}$ is properly trained, we only keep a 10\% subset of $D$ denoted as $D^{sub}$ in the server along with the well-trained model $\mathcal{M}$. When a user requests to forget a few categories of documents, the stored data $D^{sub}$ can be divided into retain set $D^{sub}_{r}$ and forget set $D^{sub}_{f}$. In this paper, we will explore unlearning the request document categories for document classification in two scenarios: firstly, where both the retain set $D^{sub}_{r}$ and forget set $D^{sub}_{f}$ are employed to update model $\mathcal{M}$; secondly, where the forget set $D^{sub}_{f}$ is eliminated immediately upon user request and only the retain set $D^{sub}_{r}$ is utilized to update the model $\mathcal{M}$. For the second scenario, we will generate synthetic samples $\overline{D}^{sub}_{f}$ using only the frozen-weight model $\mathcal{M}$ to replace the usage of real forget set $D^{sub}_{f}$. The generation process is addressed in Sec.~\ref{sec:synthesis} and the experimental results for the two scenarios, w/ and w/o access to the real forget set, are discussed in Sec.~\ref{sec:scenario1} and Sec.~\ref{sec:scenario2}.

\subsection{Document Classification Model}
\begin{figure}
    \centering
    \includegraphics[width = 0.9\linewidth]{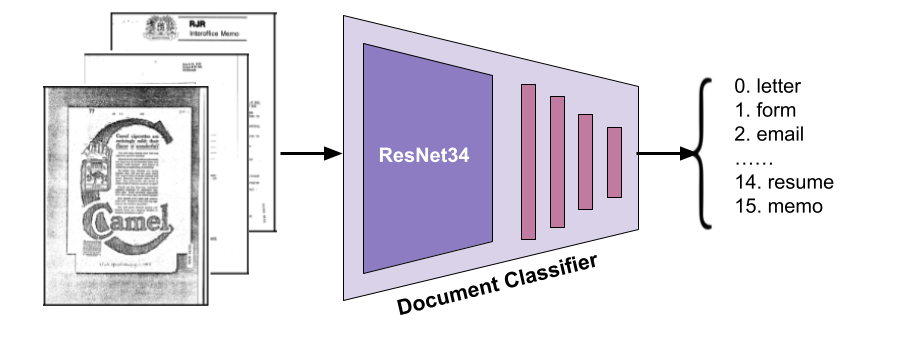}
    \caption{Architecture of the proposed backbone document classifier. Our model processes 224 $\times$ 224 images to classify them, the output is a probability distribution over the 16 document classes of the RVL-CDIP dataset.}
    \label{fig:arch}
\end{figure}

Our goal in this paper is to investigate how to efficiently unlearn the target data from the well-trained model, so we do not pursue the best accuracy on document classification in comparison with the state of the art. Thus, we take the popular neural network architecture ResNet34~\cite{he2016deep} as our backbone model. We initialize the ResNet34 model using the pre-trained weights on ImageNet-1k~\cite{deng2009imagenet}, and then train with the full training set of RVL-CDIP dataset~\cite{harley2015evaluation}. We also apply Kaiming normalization~\cite{he2015delving} on each convolutional layer and add several linear layers for final prediction. The architecture of our backbone model is shown in \cref{fig:arch}.

\subsection{Machine Unlearning Baselines}

Machine unlearning methods can be categorized into Exact Unlearning (EU) and Approximate Unlearning (AU). EU does not retain any knowledge of the removed target data, and one straightforward approach is to retrain from scratch using only retain set. This approach represents the upper bound of accuracy performance. However, it entails a significant computational and time consumption. AU takes into account the trade-off between accuracy and efficiency. It initiates with well-trained models, aiming to forget specific categories upon user request while preserving high performance on retain data promptly.

For document category forgetting, we introduce in this work, AU machine unlearning techniques based on the \textbf{Random Label (RL)} strategy. RL consists of creating a new dataset by randomly assigning new labels to the forget set $D^{sub}_{f}$ samples, ensuring that the labels in $D^{sub}_{f}$ do not reappear in the new forget set, then merging the randomly labeled forget set with the retain set $D^{sub}_{r}$ to update the model $\mathcal{M}$. RL is to directly ruin the knowledge of forget set by assigning random labels. The important thing for random labeling is to maintain the high performance for retain set. 

Moreover, two baselines are used for reference: \textbf{Retrain from scratch (RT)} and  \textbf{Fine-tune (FT)}. In RT, an EU approach, upon user request, the well-trained model $\mathcal{M}$ is eliminated immediately, and a new model is retrained from scratch with the retain set $D^{sub}_{r}$. In FT, an AU approach, the well-trained model $\mathcal{M}$ is fine-tuned on the retain set $D^{sub}_{r}$ only. Given the finding from the work~\cite{mccloskey1989catastrophic} that, when training on new tasks or categories, a neural network tends to forget the information learned in the previous trained tasks. FT is a method to take advantage of the finding above, aiming at forgetting the target category inherently. Since we do not shift to a totally new task for the FT model, the knowledge corresponding to the forget set is not guaranteed to be eliminated or overlaid by know knowledge. It's worth mentioning that both RT and FT rely solely rely on the retain set without requiring access to the forget set.

\subsection{Subset Selection Strategy}
Given our interest in investigating machine unlearning for document classification within a resource-limited environment, particularly one where only 10\% of the training data is retained on the server, we examine four subset selection strategies to assess their effects on the performance of both retained and forgotten categories:
\begin{itemize}
    \item \textbf{Random selection}. Random select 10\% of data from each category of training set.
    \item \textbf{Top selection}. Rank samples of each category by softmax probabilities from the well-trained model $\mathcal{M}$, and then select the top 10\% from each.
    \item \textbf{Bottom selection}. The same ranking as in the top selection strategy, and then select the bottom 10\% from each category of the training set.
    \item \textbf{Mix selection}. The same ranking as above, and then select the top 5\% and bottom 5\% from each.
\end{itemize}

\subsection{Label-guided Sample Generator}
\label{sec:synthesis}
\begin{figure}
    \centering
    \includegraphics[width = 0.98\linewidth]{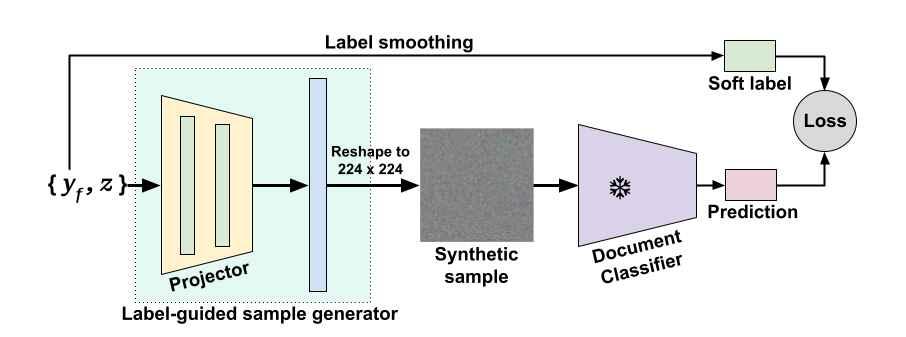}
    \caption{Architecture of label-guided sample generation method. We train the projector model to project the one-hot vector of class labels $y$ and random noise vectors $z$ into synthetic samples that can then be used to replace the need of real data for machine unlearning.}
    \label{fig:data_synth}
\end{figure}

To equip our framework capacity to proceed with the immediate removal of requested data, we propose a label-guided sample generator to produce synthetic samples $\overline{x}_f \in \overline{D}^{sub}_{f}$ replacing the usage of real forget set ${D}^{sub}_{f}$. The architecture is depicted in \cref{fig:data_synth}. The model takes the one-hot vector of class $y_i$ and a random vector $z$ as input by concatenation. These are passed to a projector module consisting of two linear layers. After that, the resultant tensor is reshaped to match the training document image size and fed to the frozen-weight well-trained document classification model (ResNet). A cross-entropy loss is then computed between the logits outputs of the ResNet and ground truth soft labels. The ground truth soft labels are simply created by applying label smoothing \cite{muller2019does} on the one-hot encoding of the targets. After training, the projector is used to generate a synthetic forget data $\overline{D}^{sub}_{f}$ that will be used with the real retain data  $D^{sub}_{r}$  for the unlearning process.



\section{Experiments}
We show the performance of our proposed methods to unlearn a single class of document dataset across a variety of settings. To make a clear comparable, all the experiments are conducted to forget the samples of the first category "letter" from the well-trained model. The overall results are shown in classification accuracy. We use $ACC_{r}$ and $ACC_{f}$ to represent the accuracy on retain and forget sets within the training set, respectively. Similarly, $ACC^{'}_{r}$ and $ACC^{'}_{f}$ denote the accuracy on retain and forget sets within the test set, respectively. We aim for higher accuracy for the retain set ($ACC_{r}$ and $ACC^{'}_{r}$), and conversely, lower accuracy for the forget set ($ACC_{f}$ and $ACC^{'}_{f}$). The expected directions are denoted by $\uparrow$ and $\downarrow$. 

\subsection{Dataset and implementation details}
The RVL-CDIP dataset~\cite{harley2015evaluation} has been the de facto benchmark for evaluating state-of-the-art document classification methods. It consists of 400,000 grayscale document images across 16 document categories of letter, memo, email, file-folder, form, handwritten, invoice, advertisement, budget, news article, presentation, scientific publication, questionnaire, resume, scientific report, and specification, with 25,000 document images per class. There are 320,000, 40,000 and 40,000 images for training, validation and test sets, respectively. To be fairly comparable in our machine unlearning setting, we focus to forget the documents of the first category ``letter". 

All the document images are resized into 224 $\times$ 224 pixels for both training and evaluation. The baseline model $\mathcal{M}$ consists of a ResNet34 and four linear layers, each employing ReLU activations and a Dropout rate of 0.5. These linear layers have neuron counts of 256, 128, 64, and 16, respectively. The baseline model $\mathcal{M}$ is trained on the whole training set of RVL-CDIP dataset with batch size 1024 and early stop of 10 epochs on validation set. More details can be found in our code.

\begin{table}[t!h]
    \caption{Unlearning results using 100\% training set. Original results are obtained with well-trained model $\mathcal{M}$ on all categories. The unlearning methods are Retrain (RT), Fine-tune (FT), and Random Label (RL).}
    \vspace{-0.1cm}
    \label{tab:full}
    \centering
    \small
    \begin{tabular}{ccccc}
    \toprule
    \textbf{Items} & \textbf{Original} & \textbf{RT} & \textbf{FT} & \textbf{RL} \\
    \midrule
    $ACC_{r}\uparrow$ & 93.53 & 94.64 & 93.95 & \textbf{94.71} \\
    $ACC_{f}\downarrow$ & 90.64 & 0 & 0 & 0 \\
    $ACC^{'}_{r}\uparrow$ & 84.29 & 85.01 & 84.37 & \textbf{85.41} \\
    $ACC^{'}_{f}\downarrow$ & 79.10 & 0 & 0 & 0\\
    \multirow{2}{*}{Epochs/Iters} & 40 epoch & 40 epoch & 3.2 epochs & \textbf{2.56 epochs}\\
     & (12500 iters) & (12500 iters) & (1000 iters) & \textbf{(800 iters)}\\
    \bottomrule
    \end{tabular}
    \vspace{-0.2cm}
\end{table}

\subsection{Unlimited training iterations}
 Following the most recent machine unlearning methods~\cite{lin2023erm,chen2023boundary}, we make use of 100\% of training set to do unlearning in an unlimited training iterations setting as shown in Tab.~\ref{tab:full}. Here we can find that all the three unlearning methods, retraining, fine-tuning and random labeling, can achieve satisfactory performance while making accuracy on forget set 0 and maintaining high accuracy on retain set. This phenomenon differs from the state of the art of machine unlearning methods for general image classification~\cite{lin2023erm,chen2023boundary}. Furthermore, obtaining the similar accuracy, the required training epochs are different. Retraining from scratch takes the same number of epochs as training the original model $\mathcal{M}$. Fine-tuning and random-labeling spend less than one-tenth training epochs to obtain the same accuracy.

\subsection{Restricted unlearning scenario}
\label{sec:scenario1}
In the realistic use cases, the AI service provider cannot store all the training set from users due to the limitation of storage and the privacy issues. We restrict to store only a subset of 10\% of training set for future unlearning. Additionally, we limit the following experiments to only 300 training iterations to ensure a fair comparison and prevent consistently reaching 0 accuracy on the forget set. The results are shown in Tab.~\ref{tab:select}. From the table, we can conclude several findings. First, RT method can quickly reach better performance with top selection strategy comparing to other subset selection strategies, because the top samples contain the dominant information for classification easily. On contrary, the bottom samples behave the opposite. Second, FT method can leverage the bottom samples to achieve the best performance comparing to other subset selection strategies, because the bottom samples lie in the boundaries among each category in the feature space that are easily to be misclassified. Based on the well-trained model, fine-tuning on the hard samples of each retain categories leads to a quick separation between each other. Third, RL method is a good trade-off between accuracy and efficiency, which achieves high accuracy on retain set and low accuracy on forget set quickly. Top samples contribute to best performance of accuracy on forget set, because the dominant features are leveraged for effective feature shifting. Conversely, bottom samples behave badly with random labeling method to obtain the worst accuracy on forget set, because bottom samples are easily to be misclassified and assigning them with random labels does not change much. Mix selection strategy can take advantage of both top and bottom samples, achieving the best performance on retain set and second-best performance on forget set using the random labeling method.

\begin{table}[t!h]
    \caption{Unlearning results using 10\% training data usage for 300 iterations. The unlearning methods are Retrain (RT), Fine-tune (FT), and Random Label (RL) for four different subset selection strategies, i.e. random selection, top selection, bottom selection, and mix selection, respectively. The most favorable results are displayed in \textbf{bold}, while the second-best results are displayed with \underline{underline}.}
    \vspace{-0.1cm}
    \label{tab:select}
    \centering
    \small
    \begin{tabular}{ccccccc}
    \toprule
     \textbf{Method} & \textbf{Metric} & \textbf{Original*} & \textbf{Random Select.} & \textbf{Top Select.} & \textbf{Bottom Select.} & \textbf{Mix Select.}\\
    \midrule
    \multirow{4}{*}{\textbf{RT}} & $ACC_{r}\uparrow$  & 93.53 & 43.85 & \textbf{53.11} & 23.96 & \underline{45.13}\\
    & $ACC_{f}\downarrow$ & 90.64 & 0 &  0 & 0  & 0 \\
    & $ACC^{'}_{r}\uparrow$ & 84.29 & 43.95 & \textbf{53.07} & 23.57 & \underline{44.99}\\
    & $ACC^{'}_{f}\downarrow$ & 79.10 & 0 & 0 & 0 & 0 \\

    \midrule
    \multirow{4}{*}{\textbf{FT}} & $ACC_{r}\uparrow$  & 93.53 & 85.32 & 86.89 & \textbf{95.13} & \underline{88.92}\\
    & $ACC_{f}\downarrow$ & 90.64 &  55.41 & \underline{38.63} & \textbf{34.27} & 41.70 \\
    & $ACC^{'}_{r}\uparrow$ & 84.29 & 78.24 & \underline{81.91} & \textbf{83.83} & 79.70\\
    & $ACC^{'}_{f}\downarrow$ & 79.10 & 48.86 & \underline{36.77} & \textbf{29.46} & 38.15 \\

    \midrule
    \multirow{4}{*}{\textbf{RL}} & $ACC_{r}\uparrow$  & 93.53 & 89.20 & 79.26 & \underline{89.44} &  \textbf{91.15}\\
    & $ACC_{f}\downarrow$ & 90.64 & 0.46 &  \textbf{0.14} &  0.70 &  \underline{0.32} \\
    & $ACC^{'}_{r}\uparrow$ & 84.29 & \underline{81.27} & 75.22 &  79.12 & \textbf{81.31}\\
    & $ACC^{'}_{f}\downarrow$ & 79.10 & 0.24 & \textbf{0.12} &  0.45 &  \underline{0.20} \\
    \bottomrule
    \end{tabular}
    \vspace{-0.2cm}
\end{table}


\begin{figure}[ht!]
\minipage{0.45\textwidth}
  \includegraphics[width=\linewidth]{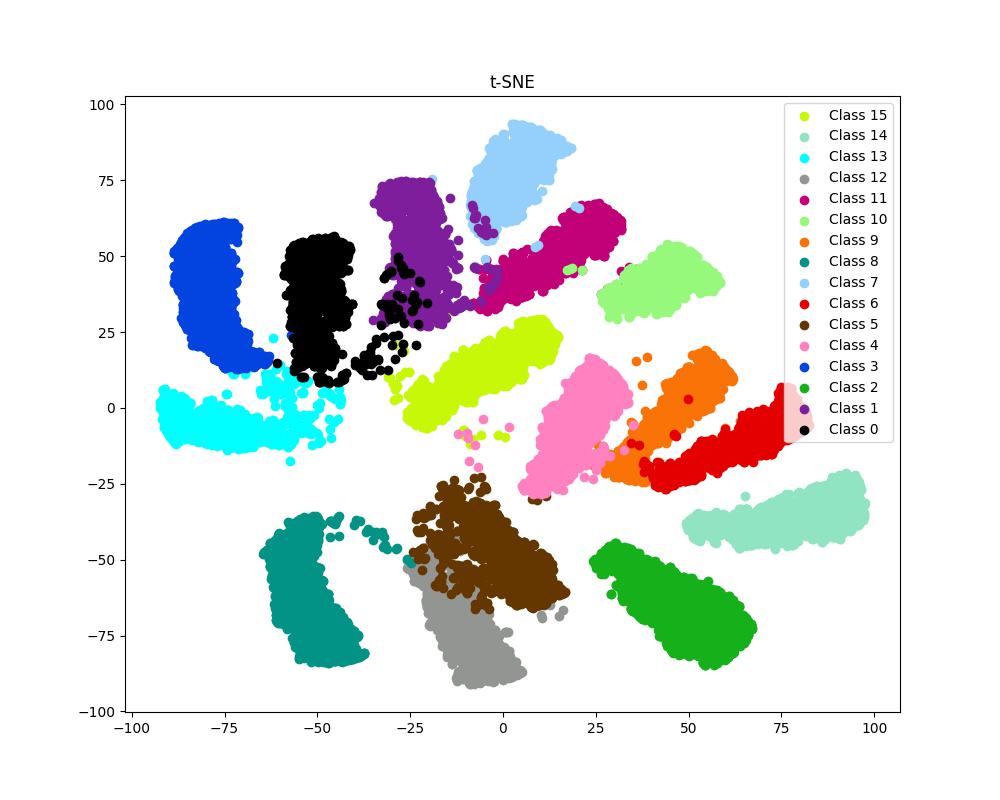}
  \vspace{-0.85cm}
  \caption{Original.}\label{fig:original}
\endminipage\hfill
\minipage{0.45\textwidth}
  \includegraphics[width=\linewidth]{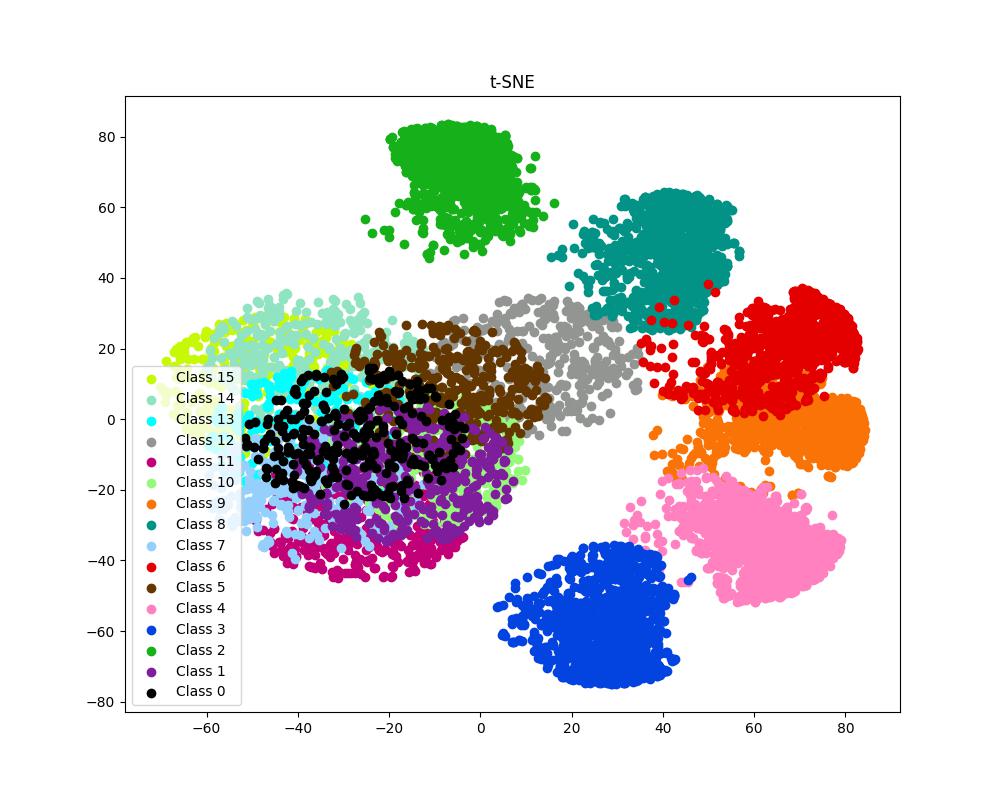}
  \vspace{-0.85cm}
  \caption{Retain (RT).}\label{fig:retrain}
\endminipage \\
\minipage{0.45\textwidth}
  \includegraphics[width=\linewidth]{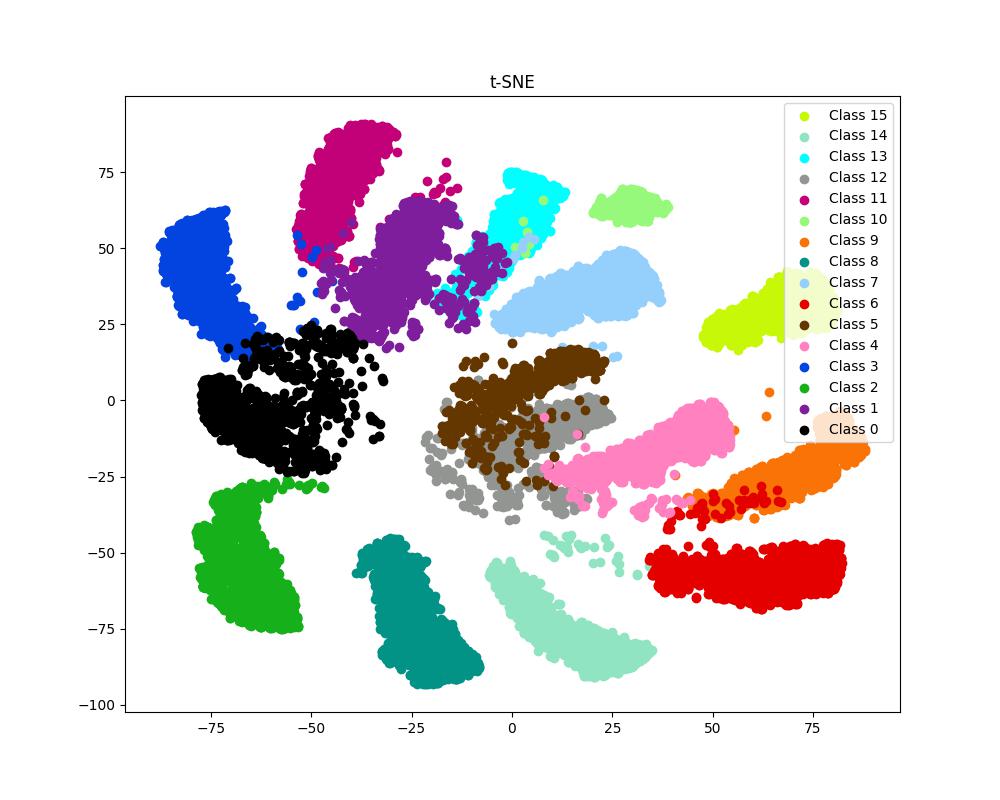}
  \vspace{-0.85cm}
  \caption{Fine-tune (FT).}\label{fig:finetune}
\endminipage\hfill
\minipage{0.45\textwidth}%
  \includegraphics[width=\linewidth]{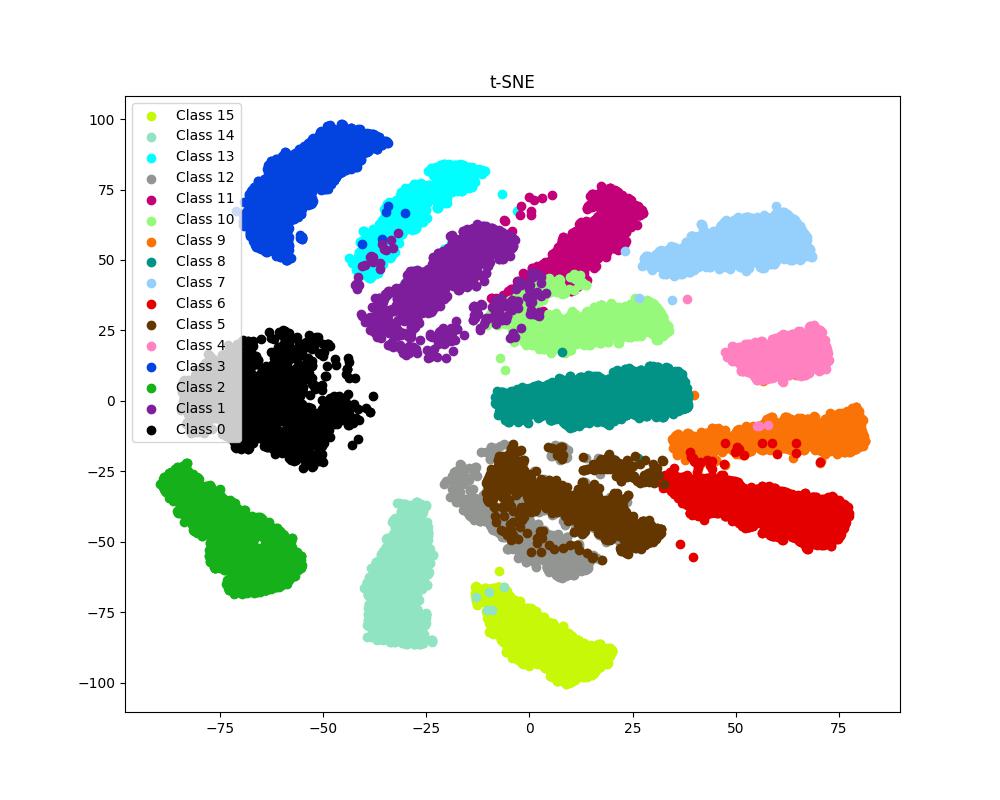}
  \vspace{-0.85cm}
  \caption{Random Label (RL).}\label{fig:randomlabel}
\endminipage
\end{figure}

We utilize t-SNE to visualize the last layer features of the proposed document classifier for the original well-trained model (\cref{fig:original}) and the three unlearning methods, retraining from scratch (\cref{fig:retrain}), fine-tuning (\cref{fig:finetune}), and random labeling (\cref{fig:randomlabel}), under mix selection scenario. In all the t-SNE visualizations, each document category corresponds to a specific color, while the forget category 0 "letter" is depicted always in black. For the original well-trained model in~\cref{fig:original}, it is evident that the majority of categories exhibit distinct boundaries, with minimal overlap between them. For the retraining method shown in~\cref{fig:retrain}, since forget set is not seen by the model, the forget samples are projected overlaid onto other categories, achieving 0 accuracy. But it is worth to note that the forget samples are still clustered together even without training. During training, the document classifier gradually develops clustering capability and assigns each category a specific label. Even if one category is not explicitly trained, inherent consistency ensures that samples from unseen categories can still be effectively clustered in the feature space. For the fine-tuning method in~\cref{fig:finetune}, the model is reinforced with samples from the retain set, but as no new categories are introduced during training, the forget samples remain in the feature space unaffected. The drop in accuracy on the forget set occurs solely due to the phenomenon of catastrophic forgetting. For the random labeling method shown in~\cref{fig:randomlabel}, the model adapts its distribution by integrating fake category samples from the forget set with randomly assigned labels. While the consistency within each category remains, the forget samples continue to cluster together. However, the overall distribution undergoes significant change, replacing the initial feature projection area with samples from other categories. Consequently, random labeling methods can strike a favorable trade-off between accuracy and efficiency.

\subsection{Unlearning without real forget set} 
\label{sec:scenario2}
To enhance our method in a realistic scenario where users can request immediate deletion of forget data and subsequent knowledge removal from the well-trained model, we propose excluding the forget set entirely during the unlearning. We still keep 10\% subset of retain set and generate the same amount of noisy synthetic forget samples as the deleted forget set. To be fairly comparison, we also generate noisy samples randomly with the same size as document images. The results are shown in Tab.~\ref{tab:gen}. From the table we can see that, the generated samples work better than the random noise samples for both accuracy on retain and forget set. Without access to the real forget set, generated samples can even obtain a better performance than the method with real samples for accuracy on retain set, while a slightly worse yet comparable accuracy on forget set.

\begin{table}[t!h]
    \caption{Unlearning results using 10\% training data for 300 iterations. The three methods are: 10\% real forget set with mix selection strategy, random noise as forget set, and generated samples as forget set. The most favorable results are displayed in \textbf{bold}, while the second-best results are displayed with \underline{underline}.}
    \vspace{-0.1cm}
    \label{tab:gen}
    \centering
    \small
    \begin{tabular}{cccccc}
    \toprule
     \multirow{2}{*}{\textbf{Metric}} & \multirow{2}{*}{\textbf{Original*}} & \multicolumn{1}{c}{\textbf{10\% Mix real}} & \multicolumn{1}{c}{\textbf{Rand. Noise}} & \multicolumn{1}{c}{\textbf{Gen. Samp.}} \\
     & & \textbf{RL} & \textbf{RL} & \textbf{RL} \\
    \midrule
    $ACC_{r}\uparrow$ & 93.53 & \underline{91.15 $\pm$ 0.98} & 87.31 $\pm$ 2.01 & \textbf{93.85 $\pm$ 0.56}\\
    $ACC_{f}\downarrow$ & 90.64 & \textbf{0.32 $\pm$ 0.15} & 0.46 $\pm$ 0.22 & \underline{0.37 $\pm$ 0.16}\\
    $ACC^{'}_{r}\uparrow$ & 84.29 & \underline{81.31 $\pm$ 0.67} & 78.43 $\pm$ 1.51 & \textbf{83.50 $\pm$ 0.31}\\
    $ACC^{'}_{f}\downarrow$ & 79.10 & \textbf{0.20 $\pm$ 0.09} &  0.34 $\pm$ 0.16 & \underline{0.23 $\pm$ 0.12}\\
    \bottomrule
    \end{tabular}
    \vspace{-0.2cm}
\end{table}


\begin{figure}[!htb]
\minipage{0.33\textwidth}
  \includegraphics[width=\linewidth]{imgs/tsne-test-rmoutlier-thresh0.5-randomlabel-real.jpg}
  \caption{Real samples.}\label{fig:rl_real}
\endminipage\hfill
\minipage{0.33\textwidth}
  \includegraphics[width=\linewidth]{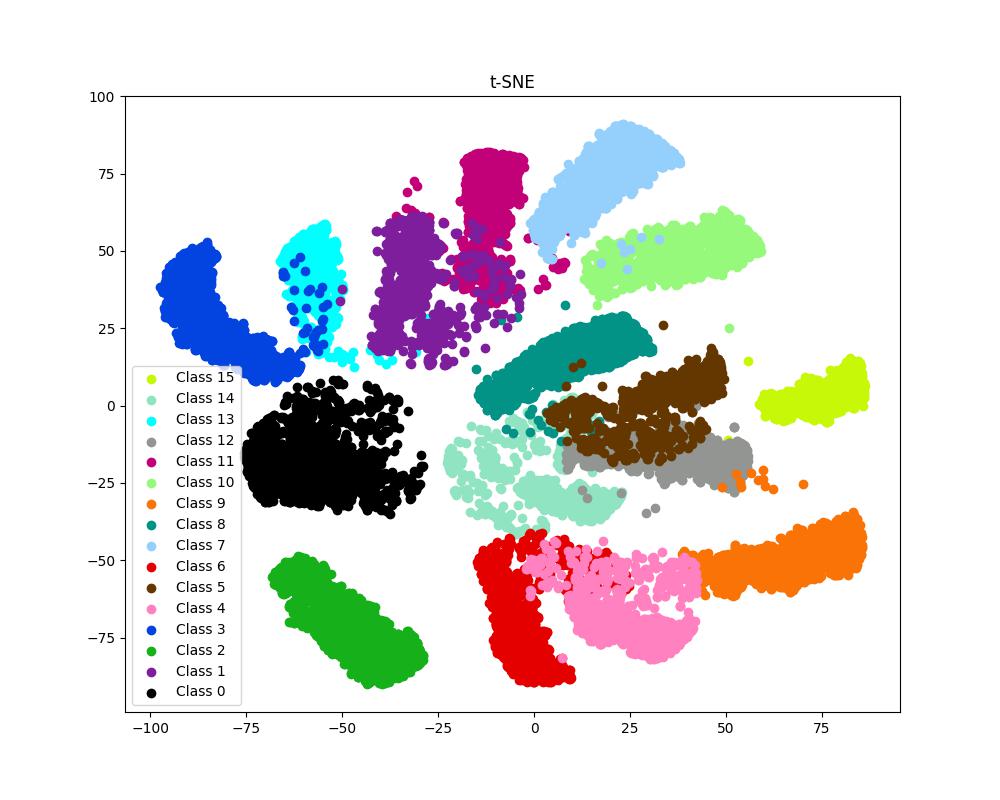}
  \caption{Random samples.}\label{fig:rl_rand}
\endminipage\hfill
\minipage{0.33\textwidth}%
  \includegraphics[width=\linewidth]{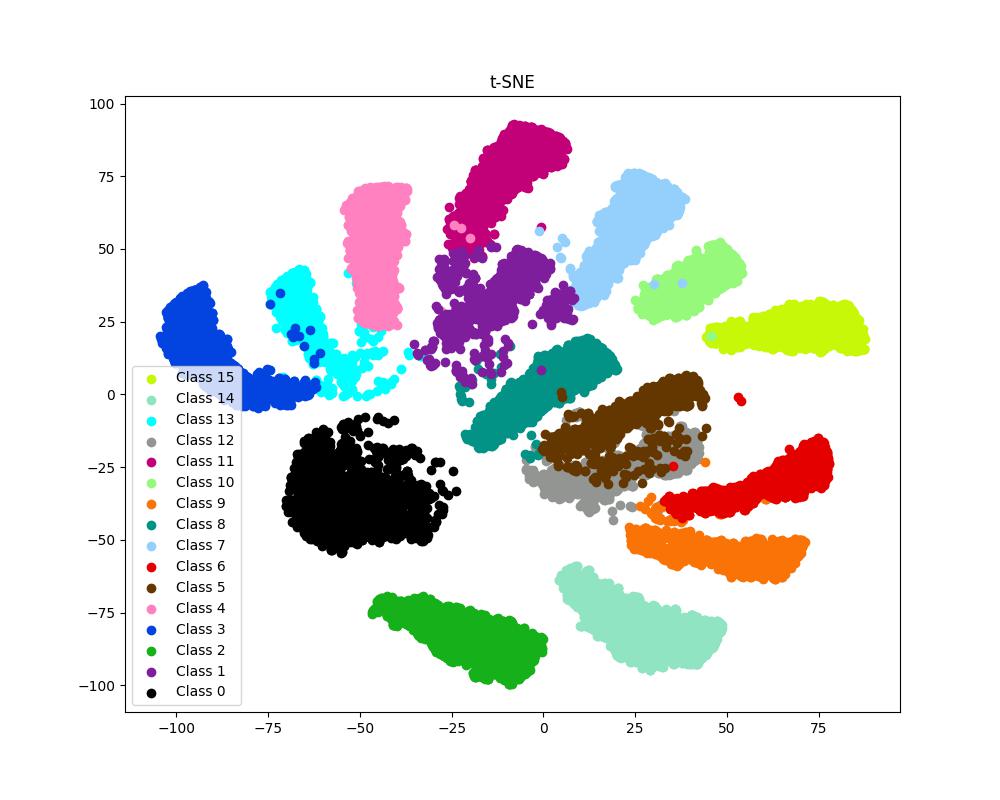}
  \caption{Generated samples.}\label{fig:rl_gen}
\endminipage
\end{figure}

Moreover, we employ t-SNE for visualizing three forget sample methods: utilizing the real forget set with the mix selection strategy (\cref{fig:rl_real}), generating random noise with the same dimensions as document images and equal in quantity to the 10\% subset of the forget set (\cref{fig:rl_rand}), and creating label-guided generated samples with dimensions matching those of document images and equivalent in quantity to the 10\% subset of the forget set (\cref{fig:rl_gen}). The random noise samples as shown in~\cref{fig:rl_rand} are independent from the the forget set, but it pushes the model to reorganize the feature distribution by adding noise to each category. Finally we can see that the generated samples (\cref{fig:rl_gen}) compel the model to restructure the feature distribution, actively attempting to displace samples from other categories onto the initial feature area occupied by the forget set. Consequently, these generated samples prove more effective, yielding superior performance compared to random noise samples and comparable performance to real forget samples.

\section{Conclusion and Future Work}

In this paper, we have undertaken pioneering research to explore machine unlearning methods for document classification, focusing particularly on constrained usage of real data and minimal training iterations. Additionally, we propose a label-guided sample generation method, enabling users to immediately eliminate the forget set on request without affecting unlearning performance. Experimental results prove its effectiveness and demonstrate comparable performance to methods utilizing real forget sets. 

Privacy issues have emerged as a prominent issue within the broader field of deep learning models. Machine unlearning is a new research line aimed at facilitating user requests for the removal of sensitive data. Document classification serves as a fundamental application within the document analysis community, more promising in terms of wide application prospects are urgent to be studied properly by researchers. In the future, our research will delve into unlearning methods for document visual question answering, with a particular focus on real-world scenarios involving business documents. Additionally, we will shift our focus from category-level to user- or even sample-level unlearning strategies.

\section*{Acknowledgements}
Beatriu de Pinós del Departament de Recerca i Universitats de la Generalitat de Catalunya (2022 BP 00256), European Lighthouse on Safe and Secure AI (ELSA) from the European Union’s Horizon Europe programme under grant agreement No 101070617, Ramon y Cajal research fellowship RYC2020-030777-I / AEI / 10.13039/501100011033.
%
%
%
\bibliographystyle{splncs04}
\bibliography{bib}

\end{document}